\pdfoutput=1

\documentclass[11pt]{article}

\usepackage[final]{acl}

\usepackage{times}
\usepackage{latexsym}

\usepackage[T1]{fontenc}

\usepackage[utf8]{inputenc}

\usepackage{microtype}

\usepackage{inconsolata}

\usepackage{graphicx}
\usepackage{array}
\usepackage{amsmath}

\title{Large Language Models Meet Legal Artificial Intelligence: A Survey}

\author{Zhitian Hou$^1$, Zihan Ye$^1$, Nanli Zeng$^2$, Tianyong Hao$^3$, Kun Zeng$^{1, \star}$ \\
        $^1$School of Computer Science and Engineering, Sun Yat-sen University \\
        $^2$China Mobile Internet Co., Ltd. \\
        $^3$School of Computer Science, South China Normal University\\
        \texttt{houzht@mail2.sysu.edu.cn;zengkun2@mail.sysu.edu.cn}
        }

\begin{document}
\maketitle
\begin{abstract}
Large Language Models (LLMs) have significantly advanced the development of Legal Artificial Intelligence (Legal AI) in recent years, enhancing the efficiency and accuracy of legal tasks. To advance research and applications of LLM-based approaches in legal domain, this paper provides a comprehensive review of 16 legal LLMs series and 47 LLM-based frameworks for legal tasks, and also gather 15 benchmarks and 29 datasets to evaluate different legal capabilities. Additionally, we analyse the challenges and discuss future directions for LLM-based approaches in the legal domain. We hope this paper provides a systematic introduction for beginners and encourages future research in this field. Resources are available at \url{https://github.com/ZhitianHou/LLMs4LegalAI}.
\end{abstract}

\section{Introduction}
Legal Artificial Intelligence (Legal AI) primarily focuses on employing AI methods to support various legal tasks, such as legal judgement prediction \citep{feng-etal-2022-legal}, legal case retrieval \citep{feng-etal-2024-legal}, and legal summarization \citep{shukla-etal-2022-legal}. Legal AI has a significant impact on the legal domain by alleviating the burden of repetitive and labor-intensive tasks faced by legal professionals. Furthermore, it provides reliable guidance to individuals with limited legal expertise, serving as an affordable form of legal assistance \citep{zhong-etal-2020-nlp}. With the success of Large Language Models (LLMs) across various domains \citep{zan-etal-2023-large, hu-etal-2024-gentranslate, kim-etal-2024-verifiner, tan-etal-2024-large}, LLMs have increasingly been applied to the legal domain to tackle a range of legal tasks \citep{wu-etal-2023-precedent, deng-etal-2024-keller, godbole-etal-2024-leveraginglongcontextlargelanguage}.

\begin{figure}[t]
    \centering
    \includegraphics[width=\columnwidth]{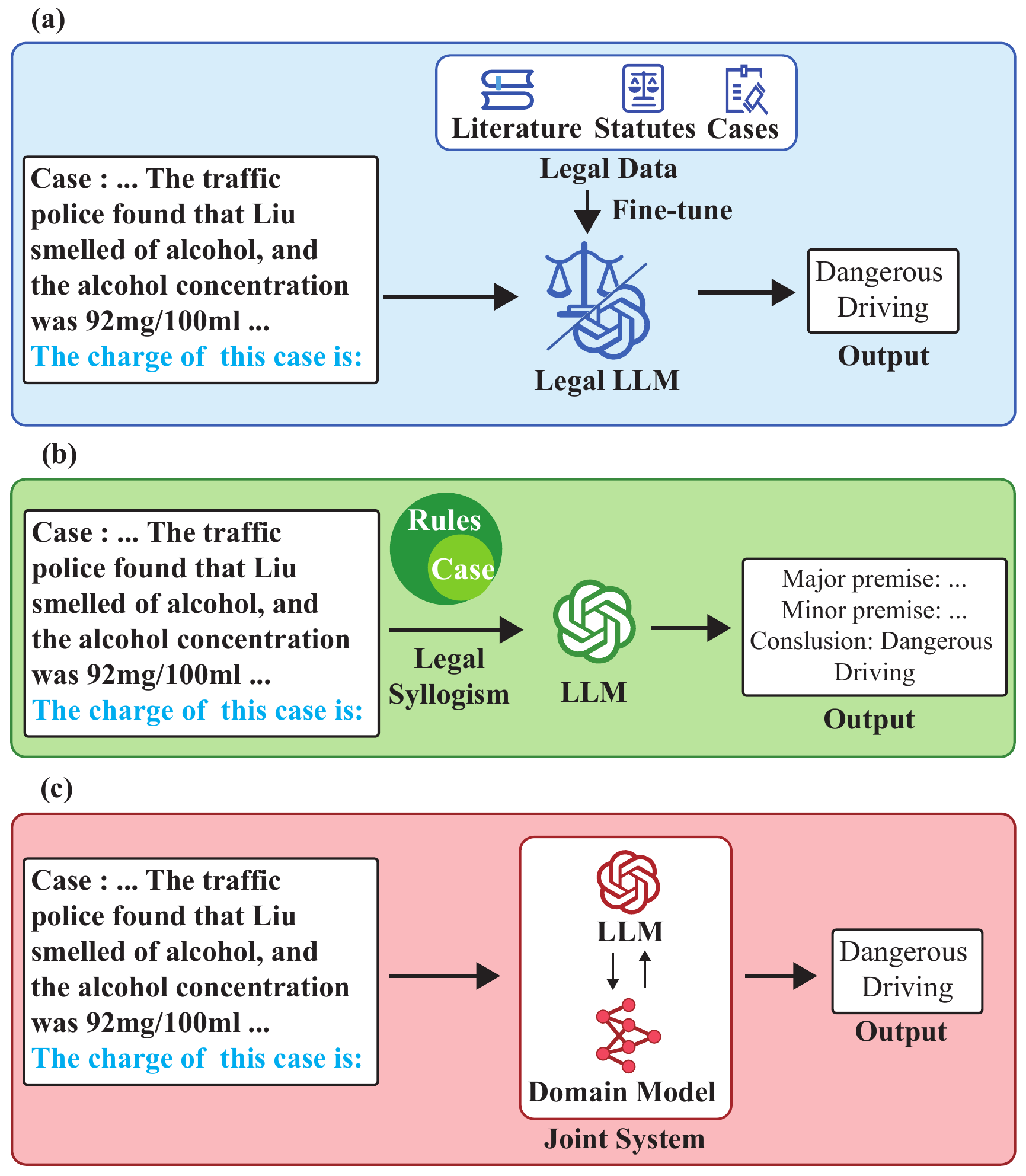}
    \caption{An example of LLMs in legal judgement prediction task. (a) is a pipeline of fine-tuning a new legal LLM. (b) and (c) are LLM-based frameworks for the task. (b) utilizes legal syllogism within the prompt. (c) uses a system jointed by LLM and Domain Model.}
    \label{fig:example}
\end{figure}

As illustrated in Figure~\ref{fig:example}, the application of LLMs in the legal domain involves fine-tuning new legal LLMs and LLM-based frameworks(i.e., leveraging existing LLMs in the framework without fine-tuning) to address traditional legal tasks. Additionally, datasets for the training and evaluation of LLMs have been developed. Despite the impressive performance of LLM-based approaches in the legal domain, a systematic review and analysis of these studies are still lacking. To fill this gap, this paper provides an in-depth exploration of the various studies for LLMs in legal domain, focusing on current datasets, advanced methods, challenges, and future directions \footnote{We discuss the related surveys and analyse the venues of papers included in this survey in Appendix~\ref{app: related survey}.}.

\begin{figure*}[t]
    \centering
    \includegraphics[width=0.9\textwidth]{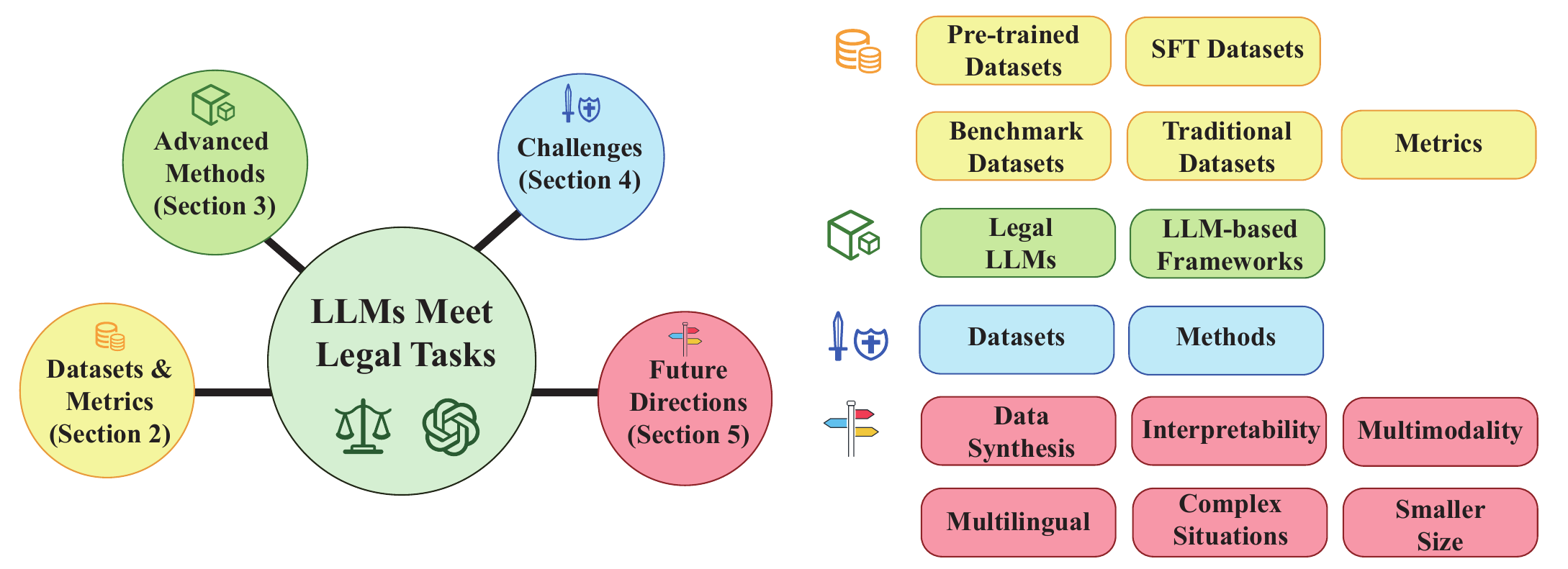}
    \caption{The organization of this survey.}
    \label{fig:organization}
\end{figure*}

In this survey, our contributions are summarized as follows: 
\begin{itemize}
    \item \textbf{Comprehensive Survey:} To the best of our knowledge, this is the first systematic review of Legal AI datasets both traditional and LLM-specific, and LLM-based approaches include legal LLMs and LLM-based frameworks.
    \item \textbf{Datasets Analysis:} We provide an analysis of existing Legal AI datasets, offering a thorough overview of their characteristics.
    \item \textbf{Meticulous Taxonomy:} We present a detailed taxonomy, distinguishing between legal LLMs and LLM-based frameworks. Furthermore, we categorize these frameworks based on the tasks they address. 
    \item \textbf{Challenges and Future Direction:} We discuss the current challenges faced by LLM-based approaches in legal applications and propose suggestions for future research.
\end{itemize}

\textbf{Survey Organization} As shown in Figure~\ref{fig:organization}, we first present datasets and metrics (\S \ref{sec: datassets}), followed by a discussion of relevant studies on advanced legal LLMs and LLM-based frameworks (\S \ref{sec: methods}). Additionally, we discuss the challenges (\S \ref{sec: challenges}) and outline future directions (\S \ref{sec: future directions}).


\section{Datasets And Metrics} \label{sec: datassets}
In this section, we review various datasets in the legal domain, including pre-training datasets, supervised fine-tuning datasets, benchmark datasets for legal LLMs, and the traditional datasets used for training and evaluating LLM-based frameworks. Additionally, we present the metrics associated with each primary task in both benchmarks and traditional datasets.

\subsection{Pre-trained Datasets}

\begin{table*}[t]
  \centering
  \begin{tabular}{m{4cm}m{1.5cm}m{2.25cm}m{6cm}}
    \hline
    \textbf{Dataset} & \textbf{Language} & \textbf{Scale} & \textbf{Text Type}\\
    \hline
    HanFei* \citeyearpar{he-etal-2023-HanFei} & Chinese & 60G & Legal Case Documents, Statutes, Litigation Documents, Legal news\\
    \hline
    JEC-QA \citeyearpar{zhong-etal-2020-jecqa} & Chinese & 26K Samples & Legal Exams\\
    \hline
    LawGPT* \citeyearpar{zhou-etal-2024-lawgpt} & Chinese & 500K Samples & Legal Case Documents\\
    \hline
    WisdomInterrogatory* \citeyearpar{wu-etal-2024-wisdomInterrogatory} & Chinese & 40G & Legal Case Documents, Legal QA Dataset\\
    \hline
    InLegalLLaMA* \citeyearpar{Ghosh-etal-2024-InLegalLLaMA} & English & 10K Samples & Legal Case Documents\\
    \hline
    NyayaAnumana \citeyearpar{nigam-etal-2025-nyayaanumana} & English & 22M Samples & Legal Case Document \\
    \hline
    Pile of Law \citeyearpar{henderson-etal-2022-pileoflaw} & English & 291.5G & Court Opinions, Contracts,
    Administrative Rules, Legislative Records\\
    \hline
    SaulLM* \citeyearpar{colombo-etal-2024-saullm54b} & English & 520B & Contracts, Court Transcripts, Statutes, Legislative Records, Legal Corpus\\
    \hline
    LBOX OPEN \citeyearpar{hwang-etal-2022-amulti} & Korean & 147K Samples & Legal Case Document \\
    \hline
  \end{tabular}
  \caption{Detailed information of legal pre-trained datasets. * indicates that this is not the name of the dataset, but the dataset used by this LLM.}
  \label{tab:pre-trained datasets}
\end{table*}

\begin{table*}[t]
  \centering
  \begin{tabular}{m{4cm}m{1.5cm}m{8.5cm}}
    \hline
    \textbf{Model} & \textbf{Language} & \textbf{Data Sources}\\
    \hline
    Lawyer LLaMA \citeyearpar{huang-etal-2023-lawyerllama} & Chinese & JEC-QA \citeyearpar{zhong-etal-2020-jecqa}\\
    \hline
    DISC-LLM \citeyearpar{yue-etal-2023-disclawllm} & Chinese & CAIL2018 \citeyearpar{xiao-etal-2018-cail}, CAIL2020 \citeyearpar{cail2020}, CJRC \citeyearpar{duan-etal-2019-CJRC}, JEC-QA \citeyearpar{zhong-etal-2020-jecqa}, JointExtraction \citeyearpar{chen-etal-2020-joint-entity}, LEVEN \citeyearpar{yao-etal-2022-leven}\\
    \hline
    InLegalLLaMA \citeyearpar{Ghosh-etal-2024-InLegalLLaMA} & English & KELM-TEKGEN \citeyearpar{agarwal-etal-2021-kelm-tekgen}, TACRED \citeyearpar{zhang-etal-2017-tacred}, Re-TACERED \citeyearpar{stoica-etal-2021-retacred}, community formus, Supreme Court of India, Supreme Court of the United Kingdom\\
    \hline
  \end{tabular}
  \caption{Detailed information of SFT datasets of three legal LLMs.}
  \label{tab:sft datasets}
\end{table*}

General LLMs are usually pre-trained on large-scale general corpora, which often lack domain-specific legal knowledge. As a result, this can lead to a limited understanding and reasoning capabilities when applied to the legal domain. Consequently, legal LLMs typically require continual pre-training based on general LLMs. This process involves using large-scale legal texts to learn rich legal knowledge. However, relying solely on legal domain data may result in catastrophic forgetting of the original general capabilities of the model. Therefore, many legal LLMs employ both general and legal pre-trained data during continual pre-training. In this paper, we focus on legal domain pre-trained datasets. We provide detailed information on the scale, language, and types of legal texts of these datasets in Table~\ref{tab:pre-trained datasets}. We observe that legal LLMs across different languages use similar types of pre-trained data, with legal case documents being the most important type for continual pre-training \citep{zhong-etal-2020-jecqa, hwang-etal-2022-amulti, he-etal-2023-HanFei, Ghosh-etal-2024-InLegalLLaMA, wu-etal-2024-wisdomInterrogatory, zhou-etal-2024-lawgpt, nigam-etal-2025-nyayaanumana}.

\subsection{Supervised Fine-Tuning Datasets} \label{sec: sft datasets}
Supervised Fine-Tuning (SFT) datasets used for developing legal LLMs are predominantly synthesized by general LLMs such as ChatGPT or GPT-4 \citep{achiam2023gpt4}. Synthetic methods include Stanford Alpaca \citep{stanford-alpaca} and Self-Instruction \citep{wang-etal-2023-self-instruct}. The sources of synthetic data can be categorized into five main types: (1) existing traditional datasets, (2) legal consultation websites, (3) legal case documents, (4) legal exams, and (5) legal QA websites. Several studies \citep{huang-etal-2023-lawyerllama, yue-etal-2023-disclawllm} utilized reference statutes to improve the quality of the synthesized datasets and mitigate hallucination issues during data generation. Detailed information on the SFT datasets used by different models is provided in Table~\ref{tab:sft datasets}, with further comprehensive details in Appendix~\ref{app: sft datasets}.

\subsection{Benchmark Datasets}
Benchmark datasets are crucial for evaluating the performance and generalizability of models. By providing standardized tasks and evaluation metrics, these datasets enable researchers to compare models, track advancements in the field, and identify areas for improvement \citep{chalkidis-etal-2022-lexglue, guha-etal-2023-legalbench, fei-etal-2024-lawbench}. Moreover, they ensure that models are evaluated across diverse legal texts, thereby enhancing their robustness and applicability. In this survey, we provide an overview of 11 benchmark datasets used to assess general legal capabilities: CaseLaw \footnote{\url{https://case.law/}}, DISC-Law-Eval \citep{yue-etal-2023-disclawllm}, KBL \citep{kimyeeun-etal-2024-developing}, LAiW \citep{dai-etal-2025-laiw}, LawBench \citep{fei-etal-2024-lawbench}, LBOX OPEN \citep{hwang-etal-2022-amulti}, LegalBench \citep{guha-etal-2023-legalbench}, LexEval \citep{li-etal-2024-lexeval}, LexGLUE \citep{chalkidis-etal-2022-lexglue}, SimuCourt \citep{he-etal-2024-agentscourt}, SMILED \citep{stern-etal-2024-smiled}, and 4 benchmarks designed for specific legal tasks: CaseSumm \citep{heddaya-etal-2024-casesumm}, CLERC \citep{hou-etal-2024-clerc}, LeDQA \citep{liu-etal-2024-ledqa}, LexSumm \citep{santosh-etal-2024-lexsumm}.

\begin{table*}[t]
  \centering
  \begin{tabular}{m{5cm}m{10cm}}
    \hline
    \textbf{Task} & \textbf{Datasets}\\
    \hline
    Retrieval & COLIEE 2023 \citeyearpar{goebel-etal-2023-coliee2023}, ELAM \citeyearpar{yu-etal-2022-elam}, LeCaRD \citeyearpar{ma-etal-2021-lecard}, LeCaRDv2 \citeyearpar{li-etal-2024-lecardv2}, ML2IR \citeyearpar{phyu-etal-2024-ml2ir} \\
    \hline
    Legal Information Extraction & InLegalNER \citeyearpar{hussain-thomas-2024-largelanguagemodelsjudicial}, JointExtraction \citeyearpar{chen-etal-2020-joint-entity} , LEVEN \citeyearpar{yao-etal-2022-leven}\\
    \hline
    Legal Judgement Prediction & CAIL2018 \citeyearpar{xiao-etal-2018-cail}, CJO22 \citeyearpar{wu-etal-2023-precedent}, ECHR \citeyearpar{chalkidis-etal-2019-neural}, FSCS \citeyearpar{niklaus-etal-2021-swiss}, MultiLJP \citeyearpar{lyu-etal-2023-multi}\\
    \hline
    Legal QA & CaseHold \citeyearpar{zheng-etal-2021-casehold}, JEC-QA \citeyearpar{zhong-etal-2020-jecqa}, LegalCQA \citeyearpar{jiang-etal-2024-hlegalki}, Legal-LFQA \citeyearpar{ujwal-etal-2024-l2fqa}, LegalQA \citeyearpar{chen-etal-2023-legalqa}, LLeQA \citeyearpar{antoine-etal-2024-interpretable}\\
    \hline
    Legal Reasoning & SARA \citeyearpar{andrew-etal-2023-cangpt-3}, COLIEE 2022 \citeyearpar{kim-etal-2023-coliee2022}, SLJA \citeyearpar{deng-etal-2023-syllogistic}\\
    \hline
    Legal Summarization & Claritin \citeyearpar{dash-etal-2019-summarizing}, CLSum \citeyearpar{liu-etal-2024-clsum}, IN-Abs \citeyearpar{shukla-etal-2022-legal}, IN-Ext \citeyearpar{shukla-etal-2022-legal}, Me-Too \citeyearpar{dash-etal-2019-summarizing}, UK-Abs \citeyearpar{shukla-etal-2022-legal}, US-Election \citeyearpar{dash-etal-2019-summarizing}\\
    \hline
  \end{tabular}
  \caption{Detailed information of traditional datasets of representative tasks.}
  \label{tab:traditional datasets}
\end{table*}

\begin{figure}[t]
    \centering
    \includegraphics[width=\columnwidth]{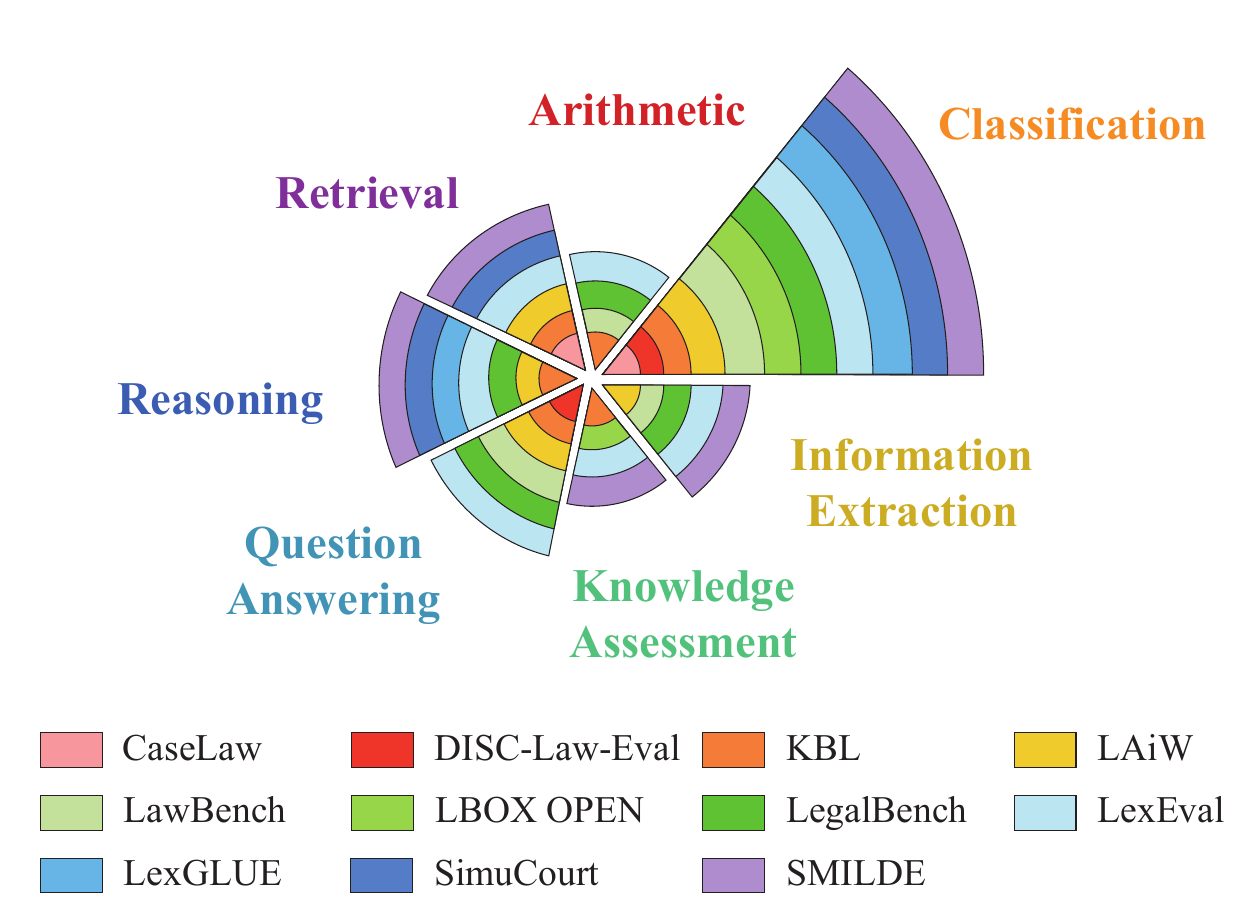}
    \caption{The capabilities assessed by benchmarks.}
    \label{fig:benchmark}
\end{figure}

\noindent \textbf{Benchmarks of General Legal Capabilities.} When developing benchmark datasets for general legal capabilities, existing traditional datasets \citep{xiao-etal-2018-cail, dan-etal-2021-cuad, yao-etal-2022-leven, wang-etal-2023-maud} are typically reformatted to align with the LLM paradigm, often adopting a QA format \citep{guha-etal-2023-legalbench, fei-etal-2024-lawbench}. 
We categorize the tasks from 10 benchmark datasets into 7 core capabilities: Arithmetic, Classification, Information Extraction, Knowledge Assessment, Question Answering, Reasoning, and Retrieval. As illustrated in Figure~\ref{fig:benchmark}, we analyse the capabilities evaluated by each benchmark. The results reveal that Classification (e.g., Clause Classification and Legal Judgement Prediction) is emphasized across all benchmarks. In contrast, Arithmetic (e.g., Crime Amount Calculation) and Knowledge Assessment (e.g., Legal Concept Memorization) are underrepresented. A detailed description of each capability can be found in Appendix~\ref{app: task definition}. 

\noindent \textbf{Benchmarks of Specific Legal Tasks.} In addition to general legal capabilities, several benchmark datasets focus specifically on evaluating performance in particular legal tasks. \textbf{CaseSumm} is a long-text summarization dataset, collecting court opinions and their official summaries from U.S. Supreme Court. \textbf{CLERC} is a dataset designed for long-context retrieval and analysis generation, using federal case law in U.S. \textbf{LeDQA} is a Chinese legal case document-based question answering dataset including 48 element-based questions. \textbf{LexSumm} is a benchmark designed to evaluate English legal summarization tasks, comprising 8 datasets from diverse jurisdictions, including the US, UK, EU, and India. 

\begin{figure*}[t]
    \centering
    \includegraphics[width=0.93\textwidth]{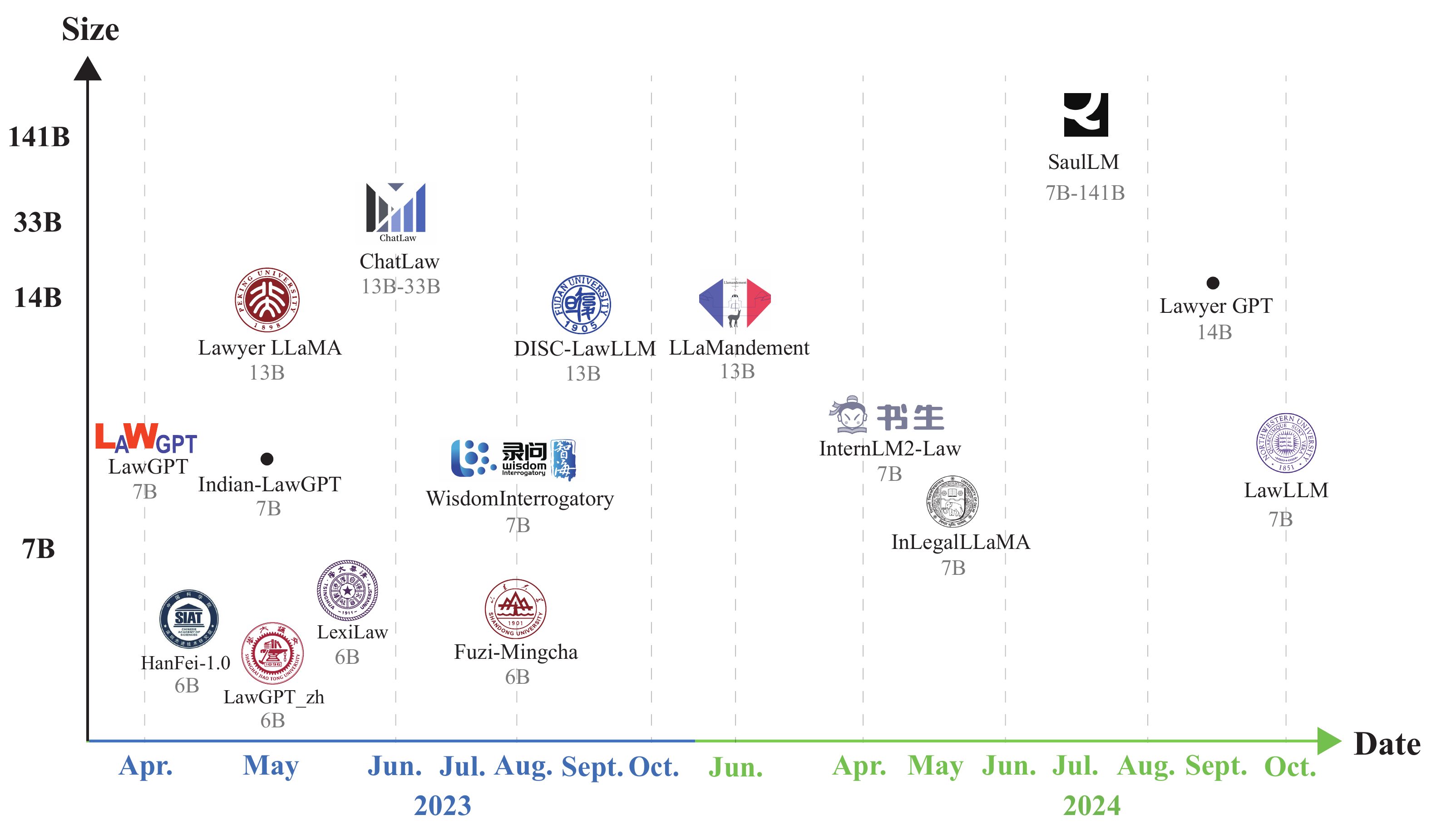}
    \caption{The timeline of Legal LLMs, with only the largest model sizes plotted for visual clarity. $\bullet$ denotes that we could not determine the logo of the LLMs.}
    \label{fig:legal llms}
\end{figure*} 

\subsection{Traditional Datasets}
In this paper, traditional datasets refer to those that were not specifically designed for LLMs. They were primarily constructed for Deep Neural Networks (DNN) and Pre-trained Language Models (PLMs). In LLM-based frameworks, LLMs are sometimes considered as assistants, where their outputs are not directly used as the final outcomes. Therefore, traditional datasets are typically utilized for LLM-based frameworks training and evaluation. In Table~\ref{tab:traditional datasets}, we categorize 29 datasets across representative 6 tasks.


\section{Advanced Methods} \label{sec: methods}
In this section, we provide an overview of legal LLMs and LLM-based frameworks.

\subsection{Legal LLMs}
We conducted a detailed survey of 16 series of legal LLMs in different languages, visualizing their release dates and sizes in Figure~\ref{fig:legal llms}. Detailed information of the language, base model, size, and release dates of these LLMs can be found in Appendix~\ref{app: legal llms}.

\textbf{ChatLaw} \citep{cui-etal-2024-chatlaw} is a series of open-source Chinese legal LLMs, comprising models such as ChatLaw-13B and ChatLaw-33B. These models are trained on an extensive dataset, including legal news, forums, and judicial interpretations. \textbf{DISC-LawLLM} \citep{yue-etal-2023-disclawllm} is a Chinese legal LLM, trained on diverse legal texts such as statutes, legal commentaries, and case databases. \textbf{Fuzi-Mingcha} \citep{sdu_fuzi_mingcha} is a model built on the ChatGLM \citep{glm2024chatglm} architecture, trained on a substantial corpus of unsupervised Chinese legal texts along with supervised legal fine-tuning data. \textbf{Hanfei} \citep{he-etal-2023-HanFei} is a 7B fully parameterized Chinese legal LLM, which supports functions such as legal question answering, multi-turn dialogue, and article generation. \textbf{InternLM2-Law} \citep{fei-etal-2025-internlm} is a large language model designed to address various legal consultations related to Chinese laws. \textbf{LawGPT} \citep{zhou-etal-2024-lawgpt} is a series of models pre-trained on large-scale Chinese legal texts and fine-tuned on legal dialogue and judicial examination datasets to enhance legal domain understanding and task execution. \textbf{LawGPT\_zh} \citep{LAWGPT-zh} is a open-source Chinese LLM, fine-tuned with LoRA. It is guided by high-quality construction datasets derived from legal articles and case studies. \textbf{Lawyer GPT} \citep{yao-etal-2024-lawyer-gpt} is a legal large model that integrates domain knowledge via an external knowledge retrieval module and demonstrates legal reasoning abilities. \textbf{Lawyer LLaMA} \citep{huang-etal-2023-lawyerllama} is a Chinese legal LLM trained on an a large-scale legal dataset, capable of offering legal advice and analyzing cases by providing relevant law articles. \textbf{LexiLaw} \citep{LexiLaw} is a Chinese legal LLM designed to provide accurate legal consultation and support for professionals, students, and general users. \textbf{WisdomInterrogatory} \citep{wu-etal-2024-wisdomInterrogatory} is trained on huge legal datasets covering a range of legal tasks in both Chinese and English. \textbf{Indian-LawGPT} \citep{indian-legalgpt} is a pioneering series of open-source LLMs specifically fine-tuned to the nuances of Indian legal knowledge. \textbf{InLegalLLaMA} \citep{Ghosh-etal-2024-InLegalLLaMA} is a set of LLMs augmented with knowledge derived from an Indian legal knowledge graph. \textbf{LawLLM} \citep{shu-etal-2024-lawllm} is a multi-task LLM designed for the US legal domain, handling Similar Case Retrieval (SCR), Precedent Case Recommendation (PCR), and Legal Judgment Prediction (LJP). \textbf{SaulLM} \citep{colombo-etal-2024-saullm7b, colombo-etal-2024-saullm54b} is a series of legal LLMs based on the Mixtral \citep{jiang-etal-2024-mixtralexperts} architecture, further pretrained on a large legal corpus. \textbf{LLaMandement} \citep{gesnouin-etal-2024-llamandement} is an LLM fine-tuned by the French government, aimed at improving the efficiency and effectiveness of parliamentary session processing through the generation of summaries.

In summary, the size of most of these legal LLMs ranges from 6B to 13B. And most of them adopt the LLaMA \citep{touvron2023llamaopenefficientfoundation} architecture as the base model. We also compare the performance of multiple legal LLMs across different benchmarks in the Appendix~\ref{app: llms results}. The results show that Fuzi-Mingcha (6B) demonstrates the best performance.

\begin{figure}[t]
    \centering
    \includegraphics[width=\columnwidth]{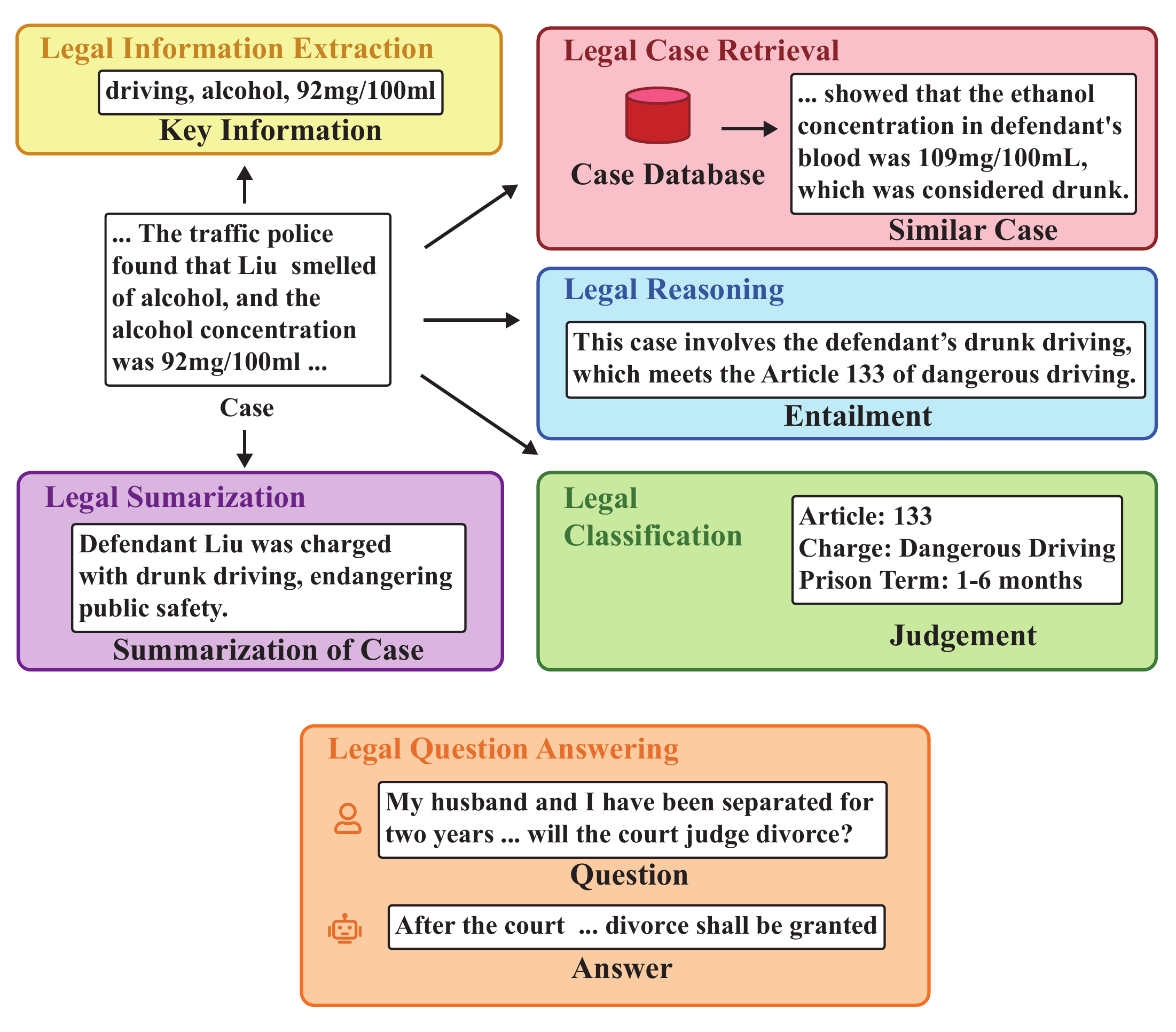}
    \caption{The examples of 6 main legal tasks.}
    \label{fig:legal tasks}
\end{figure} 

\subsection{LLM-based Frameworks}
In addition to fine-tuned legal LLMs, there are several LLM-based frameworks in legal domain. These frameworks do not directly train LLMs but instead leverage the capabilities of LLMs through other methods to accomplish various legal tasks. We classify them based on different main legal tasks, each task are visualized in Figure~\ref{fig:legal tasks}. We provide a summary of representative datasets and state-of-the-art (SOTA) methods for these tasks in Appendix~\ref{app: framework results}.

\noindent \textbf{Legal Information Extraction:} Legal information extraction aims to to extract words or phrases that have important legal significance from legal texts. The deployment of LLMs for information extraction in legal texts offers numerous potential advantages \citep{defaria-etal-2024-automatic}. Some approaches use weak supervision or sequence labeling techniques to generate annotated samples \citep{subinay-etal-2024-acasestudy}, while others transform the extraction task into a question-answering format, integrating hierarchical reasoning methods \citep{guo-etal-2024-deep}. Additionally, other frameworks leverage LLMs to extract rule pathways from legislative articles \citep{janatian-etal-2023-textstructureusinglarge} or utilize few-shot learning and prompt engineering for the task \citep{hussain-thomas-2024-largelanguagemodelsjudicial, huang-etal-2024-optimizing}. 

\noindent \textbf{Legal Judgement Prediction:} Legal Judgment Prediction is a fundamental task in legal domain. It aims to predict the judgement of the fact descriptions of legal case documents. The judgement typically consists with law articles, charges, and prison terms. Recent advancements in LJP have explored various strategies to enhance the performance with LLMs. RAG and precedent retrieval have been leveraged to integrate external legal knowledge, improving the accuracy of predictions \citep{peng-chen-2024-athena, wu-etal-2023-precedent}. Structured reasoning such as legal syllogism prompting and multi-stage frameworks refine the decision-making by systematically decomposing case facts and distinguishing confusing charges \citep{jiang-yang-2023-lot, deng-etal-2024-adapt, wang-etal-2024-legalreasoner}. Furthermore, multi-agent simulation frameworks have been introduced to improve the performance by simulating court debates, retrieving precedents, and analyzing cases \citep{he-etal-2024-agentscourt}.

\noindent \textbf{Legal Question Answering:} Legal Question Answering (LQA) seeks to address the gap between the limited number of legal professionals and the growing demand for legal assistance. \citep{nirmalie-etal-2024-cbrrag} and \citep{antoine-etal-2024-interpretable} optimize LLM output by incorporating external case-based information or utilizing retrieval pipelines. \citep{li2024bladeenhancingblackboxlarge} and \citep{jiang-etal-2024-hlegalki} focus on domain adaptation and legal knowledge integration to improve performance. \citep{moreira-etal-2024-astudy} integrate features generated by LLMs with traditional clustering to LQA. Additionally, \citep{wan-etal-2024-reformulating} use adapt-retrieve-revise pipeline to combine continual training of smaller LLMs with evidence retrieval and revision to refine answers.

\noindent \textbf{Legal Reasoning:} Legal reasoning refers to draw conclusions using methods such as chain-of-thought \citep{wei-etal-2022-cot, andrew-etal-2023-cangpt-3} based on case facts, legal statutes, judicial interpretations, and precedents. Entailment is a representative form of legal reasoning \citep{nguyen2023blackboxanalysisgptstime}. \citep{yu-etal-2022-legalpromptingteachinglanguage} found that incorporating specific legal reasoning techniques, such as the IRAC framework (Issue, Rule, Application, Conclusion), can enhance the reasoning capabilities of legal LLMs. MALR \citep{yuan-etal-2024-large} utilizes parameter-free learning to decompose complex legal tasks, while CaseGPT \citep{yang2024casegptcasereasoningframework} combines LLMs with RAG to improve reasoning. Dallma \citep{westermann2024dallma} provides a framework for semi-structured reasoning and legal argumentation. \citep{nguyen2024employinglabelmodelschatgpt} refines answers from ChatGPT using label models.

\noindent \textbf{Legal Retrieval:} Legal Retrieval aims to identify and recommend relevant legal cases or law articles based on a given legal document to support legal judgements. A prevailing paradigm of using LLM for Retrieval is the two-stage retrieval and re-ranking framework, have demonstrated significant improvements in legal case and article retrieval \citep{pham-etal-2024-aframework, nguyen-etal-2024-enhancing}. Additionally, salient information extraction and query refinement have been employed to enhance retrieval relevance, leveraging LLMs to identify and incorporate key information into retrieval models \citep{zhou-etal-2023-boosting}. Another emerging trend focuses on enhancing interpretability. Some studies have employed structured document representations and cross-matching techniques to improve the interpretability of retrieval models. \citep{deng-etal-2024-keller, ma-etal-2024-leveraginglargelanguagemodels}. Beyond monolingual systems, multilingual legal information retrieval has been explored through graph-based Retrieval-Augmented Generation (RAG), addressing legal domain-specific challenges and mitigating LLM hallucination in low-resource languages \citep{phyu-etal-2024-ml2ir}.

\noindent \textbf{Legal Summarization:} Summarization of legal texts can enable human to review and understand cases much more efficiently. Many studies have applied LLMs to this task, demonstrating potential of LLMs in this domain \citep{deroy-etal-2023-ready, pont-etal-2023-legalsummarisationllmsprodigit, godbole-etal-2024-leveraginglongcontextlargelanguage}. Several LLM-based frameworks have been proposed, including pre-processing text for better alignment with LLM training data \citep{epps-etal-2023-adapting}, multi-level extractive frameworks \citep{chhikara-etal-2025-lamsum}, and hybrid models that combine extractive and generative approaches for enhanced summary quality \citep{da-silva-etal-2024-lessons}. Additionally, data augmentation using legal knowledge have been introduced to improve the handling of long documents and limited labeled data \citep{liu-etal-2024-clsum}.

\noindent \textbf{Other Tasks:} In addition to the aforementioned well-studied tasks, some research has also been applied to tasks such as privacy policy analysis \citep{tang2023policygptautomatedanalysisprivacy, fan-etal-2024-goldcoin}, online dispute resolution \citep{westermann2023llmediatorgpt4assistedonline}, legal concept interpretation \citep{luo2025automatinglegalconceptinterpretation}, and multi-modal scenarios \citep{gao-etal-2024-adchat-tvqa, simmons2023garbageingarbageout}. Moreover, some research devoted to multi-task collaboration \citep{kim-etal-2024-legalinformationretrieval, qin-etal-2024-gear}.

\section{Challenges} \label{sec: challenges}
In this section, we analyse the existing challenges in datasets and approaches related to LLMs.

\subsection{Datasets}
In the context of LLM research in the legal domain, several challenges arise in the construction and use of datasets. 

\noindent \textbf{Data Quality.} We summarize the dataset quality issues into three aspects: 1) \textbf{Insufficient:} To prevent the leakage of legal data, courts tend to be conservative regarding the openness of judicial data and legal documents, resulting in a limited availability of legal documents \citep{LAI2024181}. It means that some data may not be learnable by LLM-related methods. 2) \textbf{Non-standard:} Legal case documents typically follow a specific structure, consist of fact description, reasoning, decisions and so on. In countries with a Statute Law system, this structure is often implicitly conveyed through text formatting. It is delineated different sections by certain standardized indicative phrases~\citep{li-etal-2023-sailer}. However, upon analysing a large corpus of legal case documents, we observed that approximately 10\% of them contain errors in these indicative phrases. As a result, segmenting the documents using rule-based approaches such as regular expressions often yields incomplete or fragmented structures. This issue is particularly pronounced in earlier documents, which we hypothesize may stem from poor-quality digitization processes during their initial conversion to electronic formats. 3) \textbf{Synthetic Data:} As mentioned in Section~\ref{sec: sft datasets}, the development of legal LLMs requires synthetic datasets created using general LLMs due to the lack of high-quality SFT datasets. However, there is no widely adopted quantifiable metrics to evaluate the quality of these SFT datasets, making it is difficult to determine their quality and whether they align with human feedback.

\noindent \textbf{Bias and Imbalance.} The existing legal datasets also face issues related to bias and imbalance. 1) \textbf{Bias:} Some datasets may contain gender or racial biases. However, there is not much discussion on how to clean biased datasets in existing methods. If AI models learn from these biased datasets, it could result in unfair decisions in practical judicial applications. 2) \textbf{Imbalance:} Civil cases are often more prevalent than criminal cases or other types of cases \cite{LAI2024181}, leading to an imbalance in datasets. This issue limits the performance of models in handling underrepresented case types. 3) \textbf{Multimodal:} In judicial practice, materials extend beyond text to include images, videos, recordings, and other modalities. However, almost datasets are only focus on text modal. 4) \textbf{Multilingual:} Most existing datasets focus on a single language rather than multilingual. Additionally, most current research has concentrated on datasets in Chinese or English, while other languages require more attention.

\subsection{Methods}
After analyzing the existing methods related to LLMs in the legal domain, we identify four key challenges: 1) \textbf{Hallucination:} Hallucination in LLMs refers to the generation of inaccurate or non-factual information. When applied to the legal domain, fluent yet erroneous outputs of LLMs could result in the dissemination of incorrect legal knowledge or mislead individuals seeking legal assistance \citep{dahl-etal-2024-large}. Ensuring the accuracy of LLM outputs remains a significant challenge. 2) \textbf{Multimodality:} As mentioned above, judicial practice involves various modalities beyond text data. However, the majority of current approaches primarily focus on text processing, neglecting the potential role of other modalities. 3) \textbf{Multilinguality:} Most existing methods focus on single language models rather than multilingual ones. This may be due to the significant legal differences between countries, which will confuse models and reduce their performance. However, multilingual LLM-based approaches faces deeper challenges due to jurisdiction-specific laws, legal traditions, and cultural contexts. Therefore, building effective multilingual LLMs-based approaches require not only access to diverse legal corpora, but also careful alignment with the legal doctrines of each jurisdiction. 4) \textbf{Interpretability:} Interpretability is critical in legal tasks. However, many studies have still not adequately explored this aspect. Providing explanations and logical reasoning during task completion is essential to meet ethical standards for use.

\section{Future Directions} \label{sec: future directions}
Building on the challenges discussed above, we outline several potential future directions for the development of legal approaches using LLMs. These directions are discussed in detail below.

\noindent \textbf{Dataset Synthesis:} It is crucial to synthesis high-quality data in the format like instructions \citep{wang-etal-2023-selfconsistency} for LLMs fine-tuning. While significant progress has been made in general dataset synthesis \citep{tan-etal-2024-large}, research on legal datasets remains limited. Future studies could focus on synthesizing legal datasets and evaluating their quality to enable more efficient fine-tuning of legal LLMs.

\noindent \textbf{Interpretability:} In the legal domain, decision-making and reasoning processes require high transparency and interpretability. Future research should focus on enhancing the interpretability of LLMs-based approaches in legal tasks, ensuring that these models not only provide accurate legal decision but also offer justifiable explanations. RAG have already been shown to reduce hallucinations in conversational tasks \citep{shuster-etal-2021-retrieval-augmentation}. It may be an important component for the interpretability of LLM-based approaches in the legal domain.

\noindent \textbf{Multimodality:} Judicial practice relies not only on text but also on multimodal data such as recordings and video evidence. However, most current legal research focuses on text processing, overlooking the potential of other modalities. Future research could explore how to combine multimodal data with text to enhance the performance of models on legal tasks adapted to real-world scenarios.

\noindent \textbf{Multilinguality:} Existing studies have mainly focused on Chinese and English legal datasets. However, the global legal system involves multiple languages and varies greatly. Future studies should focus on multilingual legal LLMs or LLM-based frameworks that address linguistic differences between legal systems and ensure broader applicability in transnational legal contexts.

\noindent \textbf{Complex Situation:} Some existing legal tasks have been simplified. For instance, samples with multiple charges or multiple defendants are often filtered out in LJP task \citeyearpar{wu-etal-2023-precedent}. However, these scenarios exist in the real world. Ignoring such cases limits the practical application of legal LLMs and LLM-based frameworks in legal tasks. The tasks with complex situation needs to be fully explored.

\noindent \textbf{Smaller Size:} The size of existing legal LLMs is mostly between 6B and 13B, while smaller models (e.g., 1.3B) have been developed in other fields \citep{Allal2023SantaCoderDR, deepseek-coder}. The smaller size of LLMs means that they can be deployed at a lower cost, which has important impact for future applications in judicial practice. Therefore, a promising future direction would be developing smaller LLMs with comparable performance to larger models.

\section{Conclusion} \label{sec: conclusion}
In this paper, we present a comprehensive review of the existing studies on legal tasks using LLMs. Specifically, we analyse 15 benchmarks and 29 datasets for evaluating various legal capabilities. Additionally, we also introduce a survey on 16 series of LLMs, as well as meticulously categorize 47 LLM-based frameworks. Finally, we highlight primary challenges and future research directions for advancing methods using LLMs in the legal domain.

\section*{Limitations} \label{sec: limitations}
Only two authors (the first two authors of this paper) were responsible for retrieval, investigation, and categorization the existing papers. This means that although we made every effort to ensure a thorough review of the published papers, some relevant works may still have been undiscovered. Additionally, human errors may occur during the categorization of papers in the survey. We have made efforts to ensure accuracy as much as possible through original descriptions and the datasets used in papers, and cross-check between authors of this paper. Therefore, while there may still be minor errors, we believe that this review is still the most comprehensive and latest review of existing datasets and methods, and it provides the analysis of existing challenges and reveals future direction in this domain.

\section*{Ethics Statement}
This survey focuses on summarizing and analyzing LLM-related datasets and methods in the legal domain. While these datasets and approaches have been widely applied in relevant research, we cannot guarantee they can apply in real-world directly. As Legal AI, particularly the datasets and methods related to LLMs, is still in an exploratory phase, there may be risks involved. Therefore, final judgment by humans will remain essential when applying these studies to judicial practice.

\bibliography{custom}

\appendix
\section{Related Survey} \label{app: related survey}
To the best of our knowledge, there are only three surveys on LLMs in the legal domain. The first one is a survey on LLMs in law. It reviewed the applications of LLMs in legal tasks, the associated legal challenges, and relevant data resources of legal domain \citep{sun2023shortsurveyviewinglarge}. However, this survey is relatively brief, covering less than 30 papers. The second paper focuses on Legal LLMs, providing a detailed description of their characteristics, summarizing the key challenges and future directions of them \citep{LAI2024181}. However, it only explores the topic of legal LLMs and does not investigate LLM-based frameworks or survey legal datasets related to LLMs. The third paper focuses on the application of natural language processing in the legal domain, providing only a short analysis and discussion of legal LLMs and LLM-based frameworks \citep{ariai2024naturallanguageprocessinglegal}. In contrast, we investigated a substantial number of papers up until 2025, covering datasets, metrics, and approaches related to LLMs. These approaches include legal LLMs fine-tuned on legal data and LLM-based frameworks that integrate LLMs into traditional task frameworks.

In the 96 works related to the datasets and approaches in this paper, 4 works are GitHub repositories, while the rest are published as papers. Among these papers, 34 are preprints, 21 were published in NLP conferences or workshops (e.g., ACL, EMNLP, and NAACL), 20 at machine learning or AI venues (e.g., NeurIPS and AAAI), and 11 were in data mining venues (e.g., SIGIR and CIKM), and 6 at other types of venues.

\section{Task Definition} \label{app: task definition}
We will give an introduce of detailed description of each capability of benchmark datasets below. We summarized all benchmark tasks and merged similar tasks.

\subsection{Arithmetic}
Arithmetic capability refers to computing time, crime amounts, and other quantifiable aspects based on relevant legal articles and legal regulations.

\noindent \textbf{Tasks:} 1) Time Calculation: Calculate time span or time according to laws and regulations in legal texts. 2) Crime Amounts Calculation: Calculate the specified crime amounts in legal texts.

\subsection{Classification}
Classification capability refers to predicting the labels assigned to legal texts, where each sample may contain multiple labels. 

\noindent \textbf{Tasks:} 1) Clause Classification: Predict the labels of each clause in legal texts. 2) Text Classification: Prediction the labels of full legal texts. 3) Legal Judgement Prediction (LJP):  Predict three sub-tasks (i.e., law articles, charges, and prison terms) based on fact descriptions of legal case documents. 

The distinction between text classification and LJP lies in the topological relationships between the three sub-tasks of LJP.

\subsection{Information Extraction}
Information extraction capability refers to extracting corresponding content from legal texts as required.

\noindent \textbf{Tasks:} 1) Argument Extraction: Extract valuable arguments and supporting evidence from legal case documents to support legal debates and case analysis. 2) Named Entity Recognition: Extract a series of words or entities that meet the requirements from the legal text. The word includes legal definitions, events, trigger words and so on. 3) Summarization: It contains two ways to generate summarization of legal texts. One is extractive summarization, this method means extracting valuable sentences from legal texts to form summaries. The other is abstractive summarization, this approach means generating summaries of legal texts with full understanding of it, not just using the sentences in the original text.

\subsection{Knowledge Assessment}
Knowledge assessment capability refers to memorizing and reciting basic knowledge in legal domain, such as cases, concepts, common sense, legal articles, legal facts, and terminologies.

\noindent \textbf{Tasks:} 1) Document Proofreading: This task involves correcting spelling, grammar, and ordering errors in sentences extracted from legal documents, producing a syntactically accurate version. 2) Legal Concept Memorization: Recite basic knowledge in legal domain, such as concepts, legal articles and so on. 3) Legal Translation: Given legal texts, translate them from one language into another.

\subsection{Question Answering}
Question answering capability refers to identifying relevant information and generating reasonable and fluent answers based on the questions. Sometimes it is also necessary to understand user intent and make responses based on multiple rounds of conversations.

\noindent \textbf{Tasks:} 1) Legal Consultations: Provide answers based on legal knowledge and offer tailored recommendations. It need to understand the intent of questions and provide context-specific legal responses. 2) Knowledge Question Answering: Answer the questions about basic legal knowledge or select the correct answer for the questions. 

\subsection{Reasoning}
Reasoning capability refers to conducting logical inferences based on multiple texts, rather than simple question answering. It typically requires the model to inference logically or determine whether two texts are entailed or contradictory.

\noindent \textbf{Tasks:} 1) Causal Reasoning: Identify whether a court's determination of causality in discrimination cases is based on statistical analysis or direct-probative evidence. 2) Contract Natural Language Inference (NLI): Determine whether a given assertion about the legal effect of a contract excerpt is supported by the content of the excerpt. 3) Multi-hop Reasoning: Deduce a conclusion through step by step, requiring the application of Chain-of-Thought (CoT) from a given premise or fact.  

\subsection{Legal Retrieval}
Retrieval capability refers to retrieving similar cases or law articles based on legal case documents. It can recommend the most similar case and the most appropriate article to assist human decision making.

\noindent \textbf{Tasks:} 1) Law Article Recommendation: Provide suitable law articles based on legal case documents. 2) Legal Case Retrieval: Retrieve the similar case in case database based on legal case documents.

\section{Supervised Fine-Tuning Datasets} \label{app: sft datasets}
We provide detailed information of the complete SFT datasets of legal LLMs in Table~\ref{tab:complete sft datasets}. It contains the language and data sources of SFT datasets used on different models.

\begin{table*}[t]
  \centering
  \begin{tabular}{m{3.5cm}m{1.5cm}m{6cm}m{3cm}}
    \hline
    \textbf{Model} & \textbf{Language} & \textbf{Data Sources} & \textbf{Strengths}\\
    \hline
    ChatLaw \citeyearpar{cui-etal-2024-chatlaw}& Chinese & no mentioned & Arithmetic, IE, KA, QA, Reasoning, Retrieval\\
    \hline
    DISC-LLM \citeyearpar{yue-etal-2023-disclawllm} & Chinese & CAIL2018 \citeyearpar{xiao-etal-2018-cail}, CAIL2020 \citeyearpar{cail2020}, CJRC \citeyearpar{duan-etal-2019-CJRC}, JEC-QA \citeyearpar{zhong-etal-2020-jecqa}, JointExtraction \citeyearpar{chen-etal-2020-joint-entity}, LEVEN \citeyearpar{yao-etal-2022-leven} & IE, KA, QA, Retrieval\\
    \hline
    Fuzi-Mingcha \citeyearpar{sdu_fuzi_mingcha} & Chinese & CAIL, China Judgements Online, LawGPT Data \citeyearpar{zhou-etal-2024-lawgpt}, Lawyer LLaMA Data \citeyearpar{huang-etal-2023-lawyerllama}, LawRefBook & Arithmetic, Classification, Reasoning\\
    \hline
    Hanfei \citeyearpar{he-etal-2023-HanFei} & Chinese & no mentioned & Arithmetic, IE, QA\\
    \hline
    InternLM-Law \citeyearpar{fei-etal-2025-internlm} & Chinese & CAIL Competition Data, LAIC Competition Data, JEC-QA \citeyearpar{zhong-etal-2020-jecqa}, LawBench \citeyearpar{fei-etal-2024-lawbench}, LEEC \citeyearpar{zongyue2023leeclegalelementextraction}, LEVEN\citeyearpar{yao-etal-2022-leven}, legal consultation websites & Classfication, IE, KA, Reasoning\\
    \hline
    LawGPT\citeyearpar{zhou-etal-2024-lawgpt} & Chinese & CrimeKgAssitant \citeyearpar{liu-crimekgassitant}, CAIL Competition Data, LAIC Competition Data, JEC-QA \citeyearpar{zhong-etal-2020-jecqa},  China Laws \& Regulations Database, QA dataset, legal consultation websites & IE\\
    \hline
    LawGPT\_zh \citeyearpar{LAWGPT-zh}& Chinese & CrimeKgAssitant \citeyearpar{liu-crimekgassitant}, China Laws \& Regulations Database & QA, Retrieval\\
    \hline
    Lawyer GPT \citeyearpar{yao-etal-2024-lawyer-gpt} & Chinese & China Judgments Online, National Judicial Examination Center & KA, Reasoning, Retrieval\\
    \hline
    Lawyer LLaMA \citeyearpar{huang-etal-2023-lawyerllama} & Chinese & JEC-QA \citeyearpar{zhong-etal-2020-jecqa} & Classification, IE \\
    \hline
    LexiLaw \citeyearpar{LexiLaw}& Chinese & LawGPT\_zh SFT Dataset, Lawyer LLaMA SFT Dataset, QA website & Classification, KA, QA, Reasoning\\
    \hline
    WisdomInterrogatory \citeyearpar{wu-etal-2024-wisdomInterrogatory} & Chinese & no mentioned & KA, Classification, QA, Retrieval\\
    \hline
    Indian-LawGPT\citeyearpar{indian-legalgpt} & English & no mentioned & Classification, KA, QA\\
    \hline
    InLegalLLaMA \citeyearpar{Ghosh-etal-2024-InLegalLLaMA} & English & KELM-TEKGEN \citeyearpar{agarwal-etal-2021-kelm-tekgen}, TACRED \citeyearpar{zhang-etal-2017-tacred}, Re-TACERED \citeyearpar{stoica-etal-2021-retacred}, community formus, Supreme Court of India, Supreme Court of the United Kingdom & Classification\\
    \hline
    SaulLM \citeyearpar{colombo-etal-2024-saullm54b} & English & no mentioned & Classification, Retrieval, Reasoning\\
    \hline
    LLaMandement \citeyearpar{gesnouin-etal-2024-llamandement} & french & SIGNALE \citeyearpar{dila_2022} & IE, Reasoning\\
    \hline
  \end{tabular}
  \caption{Detailed information of the complete SFT datasets and performance highlights about each LLM. \textbf{IE} refers to Information Extraction, \textbf{KA} refers to Knowledge Assessment, and \textbf{QA} refers to Question Answering.}
  \label{tab:complete sft datasets}
\end{table*}

\section{Detailed Information of Legal LLMs} \label{app: legal llms}
In Table~\ref{tab:detailed information of legal LLMs}, we provide detailed information of the language, base model, size, and release dates of the legal LLMs investigated in this paper.

\begin{table*}[t]
  \centering
  \begin{tabular}{lllll}
    \hline
    \textbf{Model} & \textbf{Language} & \textbf{Base Model} & \textbf{Size} & \textbf{Release Date}\\
    \hline
    ChatLaw-13B \citeyearpar{cui-etal-2024-chatlaw} & Chinese & Ziya-LLaMA-13B-v1 \citeyearpar{fengshenbang} & 13B & 2023.06\\
    ChatLaw-33B \citeyearpar{cui-etal-2024-chatlaw} & Chinese & Anima-33B \citeyearpar{airllm2023} & 33B & 2023.06\\
    \hline
    DISC-LawLLM \citeyearpar{yue-etal-2023-disclawllm} & Chinese & Baichuan-13B-Base \citeyearpar{baichuan} & 13B & 2023.09\\
    \hline
    Fuzi-Mingcha \citeyearpar{sdu_fuzi_mingcha} & Chinese & ChatGLM-6B \citeyearpar{glm2024chatglm} & 6B & 2023.08\\
    \hline
    HanFei \citeyearpar{he-etal-2023-HanFei} & Chinese & - & 7B & 2023.05\\
    \hline
    InternLM2-Law \citeyearpar{fei-etal-2025-internlm} & Chinese & InternLM2-Chat-7B \citeyearpar{cai2024internlm2} & 7B & 2024.05\\
    \hline
    LawGPT \citeyearpar{zhou-etal-2024-lawgpt} & Chinese & Chinese-LLaMA-7B \citeyearpar{chinese-llama-alpaca} & 7B & 2023.04\\
    \hline
    LawGPT\_zh \citeyearpar{LAWGPT-zh} & Chinese & ChatGLM-6B \citeyearpar{glm2024chatglm} & 6B & 2023.05\\
    \hline
    Lawyer GPT \citeyearpar{yao-etal-2024-lawyer-gpt} & Chinese & Qwen-14B-Chat \citeyearpar{qwen} & 14B & 2024.09\\
    \hline
    Lawyer LLaMA \citeyearpar{huang-etal-2023-lawyerllama} & Chinese & Chinese-LLaMA-13B \citeyearpar{chinese-llama-alpaca} & 13B & 2023.05\\
    Lawyer LLaMA 2 & Chinese & Chinese-LLaMA-13B \citeyearpar{chinese-llama-alpaca} & 13B & 2024.04\\
    \hline
    LexiLaw \citeyearpar{LexiLaw} & Chinese & ChatGLM-6B \citeyearpar{glm2024chatglm} & 6B & 2023.05\\
    \hline
    WisdomInterrogatory \citeyearpar{wu-etal-2024-wisdomInterrogatory} & Chinese & Baichuan-7B \citeyearpar{baichuan} & 7B & 2023.08\\
    \hline
    Indian-LawGPT \citeyearpar{indian-legalgpt} & English & Indian-LLaMA-7B & 7B & 2023.05\\
    \hline
    InLegalLLaMA \citeyearpar{Ghosh-etal-2024-InLegalLLaMA} & English & LlaMA-2-7B-HF \citeyearpar{touvron2023llama2openfoundation} & 7B & 2024.05\\
    \hline
    LawLLM \citeyearpar{shu-etal-2024-lawllm} & English & Gemma-7B \citeyearpar{team2023gemini} & 7B & 2024.10\\
    \hline
    SaulLM-7B \citeyearpar{colombo-etal-2024-saullm7b} & English & Mistral-7B & 7B & 2024.03\\
    SaulLM-54B \citeyearpar{colombo-etal-2024-saullm54b} & English & - & 54B & 2024.07\\
    SaulLM-141B \citeyearpar{colombo-etal-2024-saullm54b} & English & - & 141B & 2024.07\\
    \hline
    LLaMandement \citeyearpar{gesnouin-etal-2024-llamandement} & French & LLaMA-2-7B \citeyearpar{touvron2023llama2openfoundation} & 7B & 2024.01\\
    \hline
  \end{tabular}
  \caption{Detailed information of legal LLMs.}
  \label{tab:detailed information of legal LLMs}
\end{table*}

\section{Results of Different Benchmarks of LLMs} \label{app: llms results}
The benchmark evaluation system for Chinese legal LLMs is more comprehensive than that of any other language. In contrast, benchmarks for other languages, such as English and Korean, primarily evaluate general LLMs rather than legal LLMs. Table~\ref{tab:legal-benchmark} presents the performance of various legal LLMs across four widely used benchmarks. As shown in Table~\ref{tab:legal-benchmark}, Fuzi-Mingcha and Hanfei have outstanding performances. Despite their relatively small sizes (6B and 7B, respectively), both models outperform several larger LLMs across multiple benchmarks.

\begin{table*}[t]
\centering
\begin{tabular}{l|cccc}
\hline
\textbf{Model} & \textbf{DISC-Law-Eval Benchmark} & \textbf{LAiW} & \textbf{LawBench} & \textbf{LexEval} \\
\hline
ChatLaw & \underline{14.20} & 25.77 & \underline{32.76} & 17.90 \\ 
DISC-LLM & \textbf{20.25} & - & - & - \\
Fuzi\_mingcha & - & \textbf{40.62} & \textbf{33.05} & - \\
Hanfei & - & \underline{35.69} & 29.71 & \underline{23.50} \\
LawGPT & 11.73 & 22.69 & 9.91 & 12.90 \\
Lawyer LLaMA & 14.08 & 29.25 & 25.32 & 13.00 \\
LexiLaw & 12.20 & 31.31 & 28.78 & \textbf{24.60} \\ 
WisdomInterrogatory & - & 18.83 & 31.41 & 7.00 \\
\hline
\end{tabular}
\caption{Performances of legal LLMs across multiple benchmarks. For LLMs released in different sizes, we record the best performing version. Notably, both LawBench and LexEval scores are obtained under zero-shot evaluation settings. The best score is \textbf{bolded} and the second best score is \underline{underline}.}
\label{tab:legal-benchmark}
\end{table*}

\section{SOTA of Different Task} \label{app: framework results}
Table~\ref{tab:legal-nlp-sota} provides an overview of state-of-the-art(SOTA) results on each task across various datasets and languages. For each task, we list the associated dataset with its language, the sota framework, and its corresponding evaluation metrics. Different tasks in Legal AI are evaluated using various metrics to compare model performance. For generative tasks (e.g., Legal Reasoning, Legal Summarization, and Legal QA), the ROUGE score is the primary evaluation metric. Classification tasks (e.g., Legal Judgement Prediction) and information extraction tasks typically use accuracy, precision, recall, and F1 score as the metrics. Besides, precision is often used in retrieval tasks.

\begin{table*}[t]
\centering
\begin{tabular}{m{2cm}|m{2cm}m{1.5cm}m{3cm}m{5cm}}
\hline
\textbf{Task} & \textbf{Dataset} & \textbf{Language} & \textbf{SOTA} & \textbf{Results} \\
\hline
Legal Information Extraction & Self-Construct Dataset & Chinese & \citep{guo-etal-2024-deep} & 93.79@A,\quad 90.91@P,\quad 80.39@R,\quad 85.33@F \\ \cline{2-5}
& \textit{UKET}$_{\text{ori}}$ & English & \citep{defaria-etal-2024-automatic} & 97.79@A \\
\hline
Legal Judgement Prediction & CAIL2018 & Chinese & \citep{wu-etal-2023-precedent} & 87.07@A of Law Article, 94.99@A of Charge, 48.72@A of Prison Term \\ \cline{2-5}
& ECHR & English & \citep{wang-etal-2024-legalreasoner} & 0.85@F \\
\hline
Legal Question Answering & JEC-QA & Chinese & \citep{wan-etal-2024-reformulating} & 66.2@A \\ \cline{2-5}
& LegalCQA-en & English & \citep{jiang-etal-2024-hlegalki} & 21.93@BLEU$_1$ \\
\hline
Legal Reasoning & COLIEE 2021 & English & \citep{yu-etal-2022-legalpromptingteachinglanguage} & 81.48@A \\
\hline
Legal Retrieval & LeCaRD & Chinese & \citep{deng-etal-2024-keller} & 66.84@P \\ \cline{2-5}
& COLIEE 2023 & English & \citep{nguyen-etal-2024-enhancing} & 72.77@P, 87.12@R, 80.85@F2 \\ \cline{2-5}
& Self-Construct Dataset & Burmese & \citep{nigam-etal-2025-nyayaanumana} & 87.32@ROUGE L \\
\hline
Legal Summarization & Claritin & English & \citep{chhikara-etal-2025-lamsum} & 62.66@ROUGE-Lsum \\
\hline
\end{tabular}
\caption{SOTA results for legal NLP tasks across different datasets and languages.}
\label{tab:legal-nlp-sota}
\end{table*}

\end{document}